\documentclass{article}

\PassOptionsToPackage{numbers, compress}{natbib}

\usepackage[preprint]{nips_2018}

\usepackage[utf8]{inputenc} 
\usepackage[T1]{fontenc}    
\usepackage{hyperref}       
\usepackage{url}            
\usepackage{booktabs}       
\usepackage{amsfonts}       
\usepackage{nicefrac}       
\usepackage{microtype}      
\usepackage{graphicx}
\usepackage{subcaption}
\usepackage{amsmath}
\usepackage{amssymb}
\usepackage[vlined,ruled,noend]{algorithm2e}

\title{Active Learning with TensorBoard Projector}

\author{
  Francois Luus\\
  IBM Research AI\\
  \texttt{fluus@za.ibm.com} \\
  \And
  Naweed Khan\\
  IBM Research AI\\
  IBM Research \textbar~Africa\\
  \And
  Ismail Akhalwaya\\
  IBM Research AI \&\\
  CSAM at University of the Witwatersrand
}

\begin{document}

\maketitle

\begin{abstract}
An ML-based system for interactive labeling of image datasets is contributed in TensorBoard Projector to speed up image annotation performed by humans. The tool visualizes feature spaces and makes it directly editable by online integration of applied labels, and it is a system for verifying and managing machine learning data pertaining to labels. We propose realistic annotation emulation to evaluate the system design of interactive active learning, based on our improved semi-supervised extension of t-SNE dimensionality reduction. Our active learning tool can significantly increase labeling efficiency compared to uncertainty sampling, and we show that less than 100 labeling actions are typically sufficient for good classification on a variety of specialized image datasets. Our contribution is unique given that it needs to perform dimensionality reduction, feature space visualization and editing, interactive label propagation, low-complexity active learning, human perceptual modeling, annotation emulation and unsupervised feature extraction for specialized datasets in a production-quality implementation.
\end{abstract}

\section{Introduction}
An efficient labeling method is proposed for specialized image data, which is based on varying-expense unsupervised feature extraction with Exemplar-CNN followed by semi-supervised t-SNE.
Dimensionality Reduction (DR) presents the annotator with a view of the data structure to initiate the labeling process, which depends on locating combined labeling prospects through local homogeneity assessment of clusters given graphical sample depictions.
Our interactive labeling and semi-supervised extension of t-SNE is available in the newest version of TensorBoard Projector~\cite{smilkov2016embedding} and in this paper we evaluate its potential usefulness to speed up human labeling of specialized image datasets
with the objective of maximizing classifier accuracy with minimum labeling actions.
Image datasets with relatively small differences between classes are defined as specialized in this study. 
Based on our experience with the platform we realistically emulate user interactions (Algorithm 1) for large-scale testing and comparison of labeling efficiency, although no human trials are reported.

Unsupervised representation learning based on ImageNet-pretrained Exemplar-CNN~\cite{dosovitskiy2014discriminative} that has either frozen or unfrozen CNN stages is proposed.
Active learning is incorporated into an ImageNet-pretrained CNN by \cite{wang2017cost} to progressively fine-tune feature extraction with supervision. Pool-based active learning is also used to refine a CNN feature extractor and classifier in \cite{geifman2017deep} using a coreset sampling approach. Note that these require 1000s of labels or actions for active learning.
We do unsupervised transfer learning for feature extraction, but no continual semi-supervised retraining of the CNN as in~\cite{wang2017cost,geifman2017deep}. 
Human interaction is modeled with ML in~\cite{lake2015human} with similar interactive components, but
we tightly integrate DR with interaction~\cite{sacha2017visual,salazar2015interactive} and constraining supervision~\cite{bunte2012general}.
Our contributions are 1) showing empirical efficacy of unsupervised transfer learning with pretrained models on specialized images; 2) improving semi-supervised t-SNE of~\cite{mcinnes2016sstsne} with globally normalized attractions, semi-supervised repulsion, smoothed label integration, and parameter study; 3) investigating the benefit of semi-supervision in progressive labeling;
and 4) emulating annotation by modeling local homogeneity assessment with Barnes-Hut focus kernels for realistic upper-bounds.

\section{Unsupervised Transfer Learning for Specialized Images}

\begin{figure}[t!]%
\centering%
\begin{subfigure}[b]{0.44\textwidth}
\includegraphics[width=\textwidth]{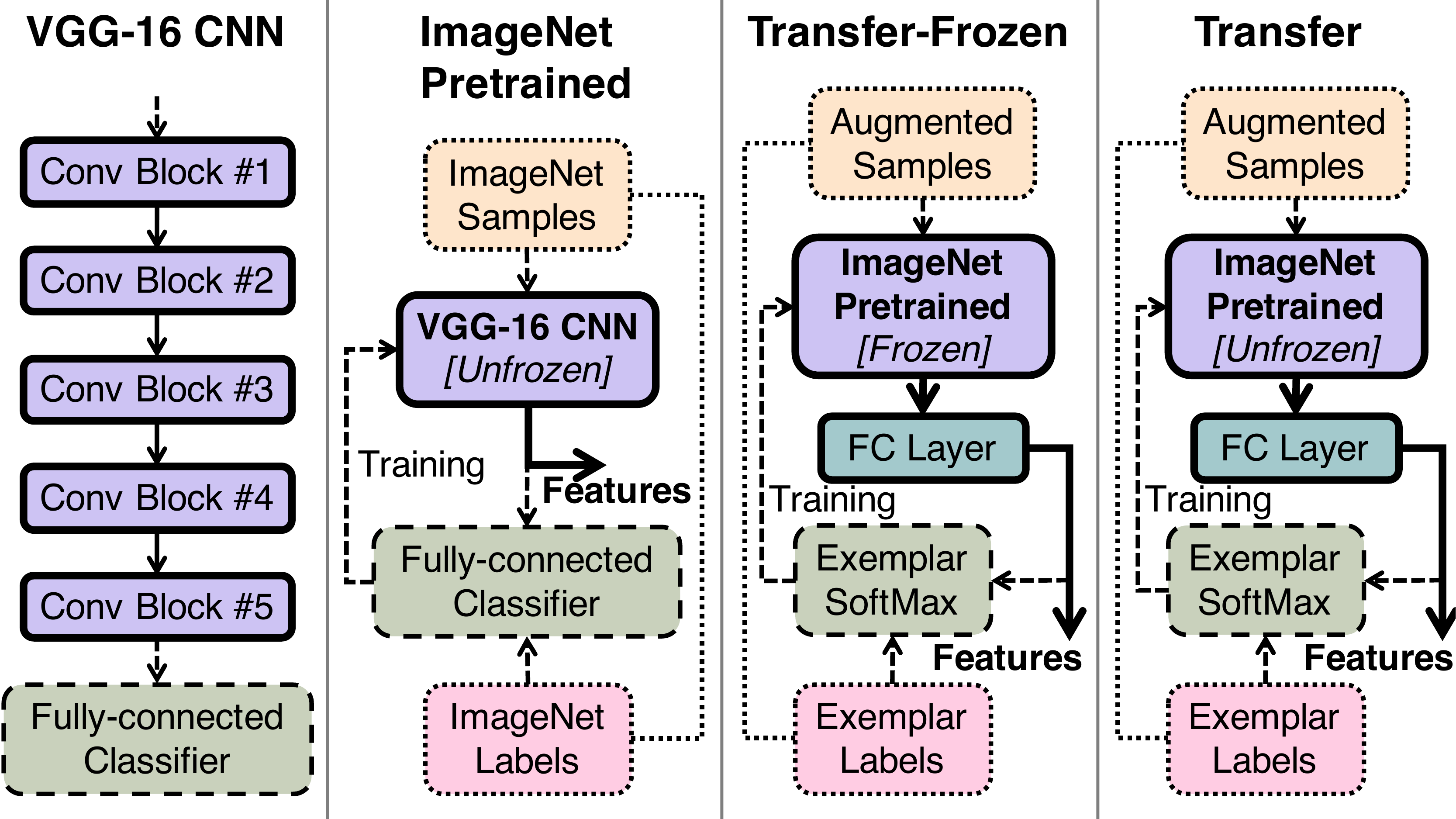}
\caption{Variable-expense feature extraction}\label{system-diagram-1}
\end{subfigure}
\begin{subfigure}[b]{0.55\textwidth}
\includegraphics[width=\textwidth]{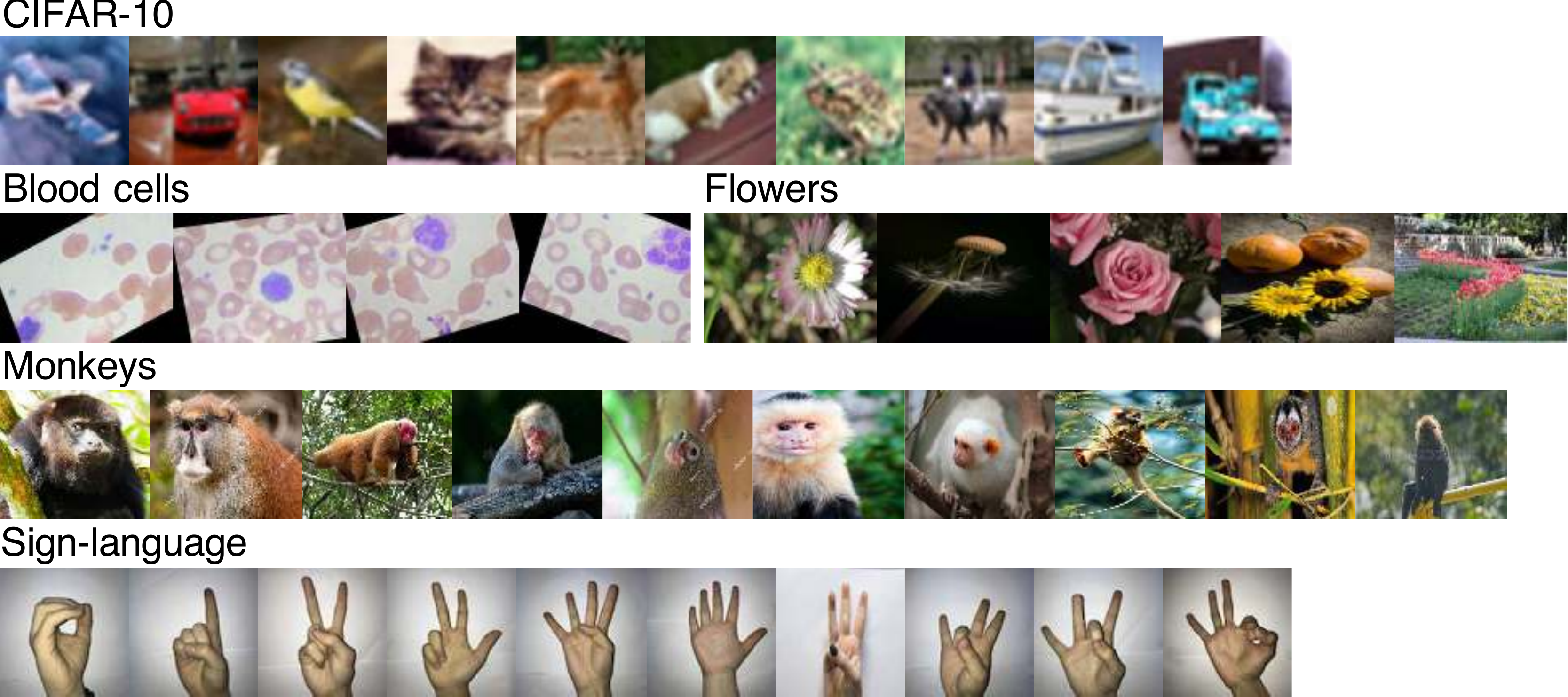}
\caption{Specialized image samples}\label{image-samples-1}
\end{subfigure}%
\caption{(a) Variable-expense feature extraction model and (b) specialized image samples.}
\end{figure}

\textbf{Rotation-Based Exemplar-CNN}: Learning features based only on predicting rotations has shown promising results in~\cite{gidaris2018unsupervised}, without the need for more extensive augmentation as in the original Exemplar-CNN~\cite{dosovitskiy2014discriminative}.
Rotation-based augmentation is used in the variable-expense feature extraction proposed in Figure~\ref{system-diagram-1}, consisting of an ImageNet-pretrained CNN~\cite{russakovsky2015imagenet}, which can be followed by a ReLU-activated fully-connected (FC) layer trained to predict exemplars of original samples. Note that groundtruth labels are not used, since each sample becomes its own surrogate class by continual augmentation through unrestricted random rotation.
The global average pooling output (dim=512) is used for our VGG-16 CNN stage~\cite{simonyan2014very}, followed by an FC layer (dim=512) for transfer learning.
CNN weights can either be frozen (Transfer-Frozen) or trainable (Transfer) during the transfer learning with Exemplar-CNN. 
The added FC layer has the same dimension as the CNN stage output and provides learning capacity for Transfer-Frozen, which should require less computational resources than the full CNN retraining required by Transfer.
\begin{table}[b!]
\centering 
\caption{4-NN classification accuracies ($\mu\pm\sigma$) in 3D t-SNE of features including pixel (Pix), preprocessed (Pre), randomly-initialized Exemplar-CNN (RN), ImageNet-pretrained (IM), Transfer-Frozen (TF), and Transfer (TR) features using VGG-16 and Inception-Resnet-v2 CNNs.}
\label{tab-inc-feats}
{\small
\begin{tabular}{@{}c@{ } @{ }c@{ } 		@{ }c@{ } 		@{ }c@{ } 		@{ }c@{ } 		@{ }c@{ }		@{ }c@{ }	}
\hline
&\multicolumn{6}{c}{\bf{VGG-16}}\\
\cline{2-7}
&  CIFAR-10 		&  Blood 		& Flowers 		& Monkey		& Signs 		&  Avg.		\\
\hline
Pix	& 24$\pm$26		&  26$\pm$22		& 38$\pm$31		& 31$\pm$28		& 62$\pm$37		&  36		\\ 
Pre	& 24$\pm$26		&  26$\pm$22		& 38$\pm$31		& 31$\pm$28		& 62$\pm$37		&  36		\\
RN & 20$\pm$23 & 42$\pm$30 & 38$\pm$30 & 28$\pm$26 & 38$\pm$34 & 33\\
IM	& \bf68$\pm$36	&  51$\pm$32		& 75$\pm$33		& 77$\pm$31		& \bf84$\pm$30	&  71		\\ 
TF	& 62$\pm$36		& \bf96$\pm$15	& 75$\pm$33 		& \bf94$\pm$20	& 76$\pm$34		&  80		\\ 
TR	& 63$\pm$36		&  95$\pm$17		& \bf77$\pm$32 	& 93$\pm$21		& 76$\pm$35		&\bf81	\\
\hline
\end{tabular}
\begin{tabular}{@{}c@{ } 		@{ }c@{ } 		@{ }c@{ } 		@{ }c@{ } 		@{ }c@{ }		@{ }c@{ }	}
\hline
\multicolumn{6}{c}{\bf{Inception-Resnet-v2}}\\
\cline{1-6}
CIFAR-10		&  Blood 		&  Flowers  		&  Monkey 		&  Signs 		&  Avg.	    	\\
\hline
24$\pm$26 		& 26$\pm$22 		& 38$\pm$31		& 31$\pm$28 		& \bf62$\pm$37 	&  36	    	\\
24$\pm$26 		& 26$\pm$22 		& 38$\pm$31		& 31$\pm$28 		& 61$\pm$37		&  36	    	\\
22$\pm$24&25$\pm$22&41$\pm$32&41$\pm$32&56$\pm$37&37\\
\bf74$\pm$35 	& 41$\pm$30 		& 70$\pm$35 		& 90$\pm$23 		& 60$\pm$38 		&  67	    	\\
74$\pm$35 		& 83$\pm$27 		& 74$\pm$34 		& \bf99$\pm$9 	& 53$\pm$38 		&  76	    	\\
73$\pm$35 		& \bf84$\pm$27 	& \bf79$\pm$33 	& 99$\pm$9 		& 53$\pm$38 		&\bf78	\\
\hline
\end{tabular}
}
\end{table}

\textbf{Specialized Images}: CIFAR-10~\cite{krizhevsky2009learning} (10 classes, 3K of 50K, 32$\times$32) is used as an example of a more general dataset in this study, to contrast against the other specialized datasets shown in Figure~\ref{image-samples-1}.
Interactive labeling depends on easy navigation of the presented feature space to discern between samples according to corresponding graphical depictions for each sample. The working size of the dataset is limited to reduce crowding and resulting sample occlusion. A maximum size of approximately 3000 (class-stratified sampling) produces a medium density feature space that can be explored and labeled.
The classification of blood cell subtypes from images have important medical applications, notably in the diagnosis of blood-based diseases, and the augmented test dataset~\cite{mooney} of the BCCD corpus~\cite{shenggan} is used (4 classes, 2400 samples, 320$\times$240).
Image classification of fauna and flora species also has wide applicability, so the monkey~\cite{monkeys} (10 classes, 1368 samples, 420$\times$360) and flower~\cite{flowers} species datasets (10 classes, 3K of 4242, 320$\times$240) are included.
The 0-9 digits expressed in sign-language~\cite{signs} (10 classes, 2050 samples, 100$\times$100) are also added, which is a more complex analogue of the MNIST dataset~\cite{lecun2010mnist}.

\begin{figure}[t!]%
\centering%
\begin{subfigure}[b]{0.199\textwidth}
\includegraphics[width=\textwidth]{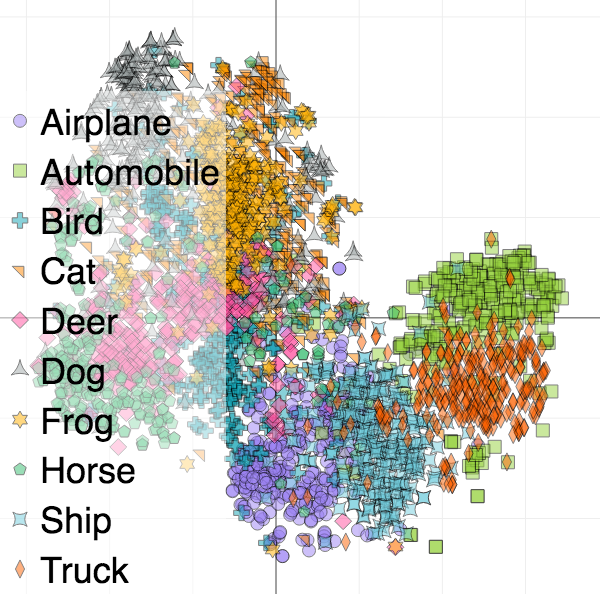}
\caption{CIFAR-10 (IM)}\label{embedding-plots-cifar-1}
\end{subfigure}%
\begin{subfigure}[b]{0.199\textwidth}
\includegraphics[width=\textwidth]{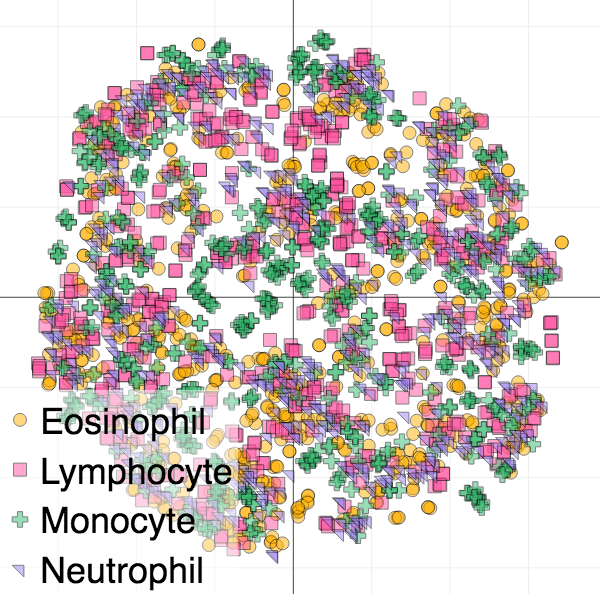}
\caption{Blood cells (IM)}\label{embedding-plots-blood-1}
\end{subfigure}%
\begin{subfigure}[b]{0.199\textwidth}
\includegraphics[width=\textwidth]{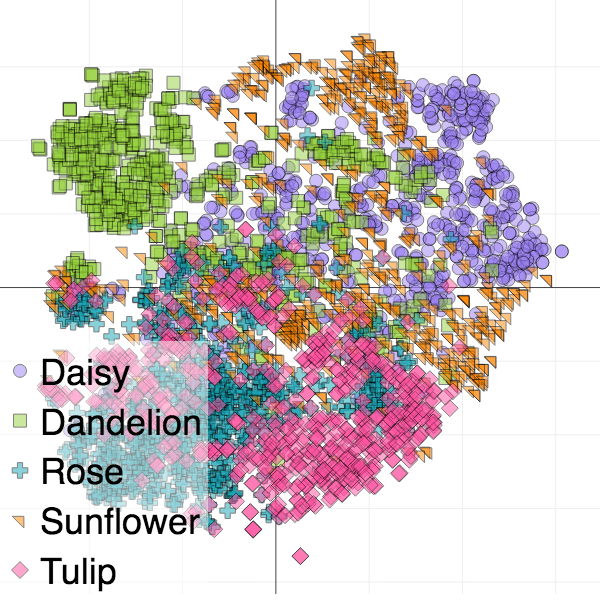}
\caption{Flowers (IM)}\label{embedding-plots-flowers-1}
\end{subfigure}%
\begin{subfigure}[b]{0.199\textwidth}
\includegraphics[width=\textwidth]{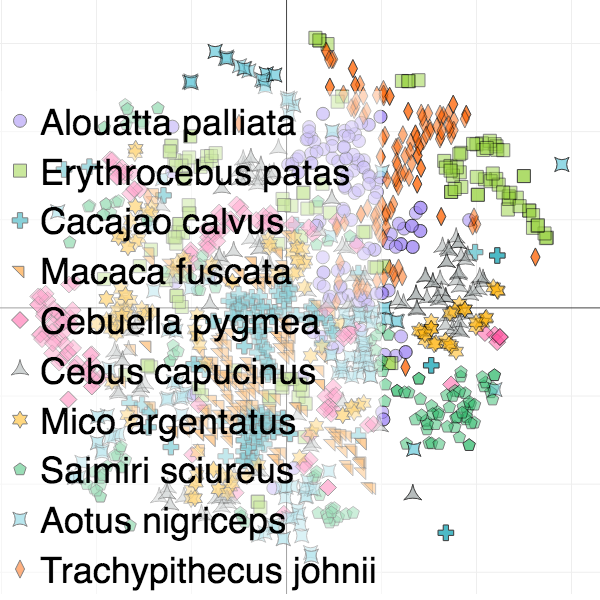}
\caption{Monkeys (IM)}\label{embedding-plots-monkey-1}
\end{subfigure}%
\begin{subfigure}[b]{0.199\textwidth}
\includegraphics[width=\textwidth]{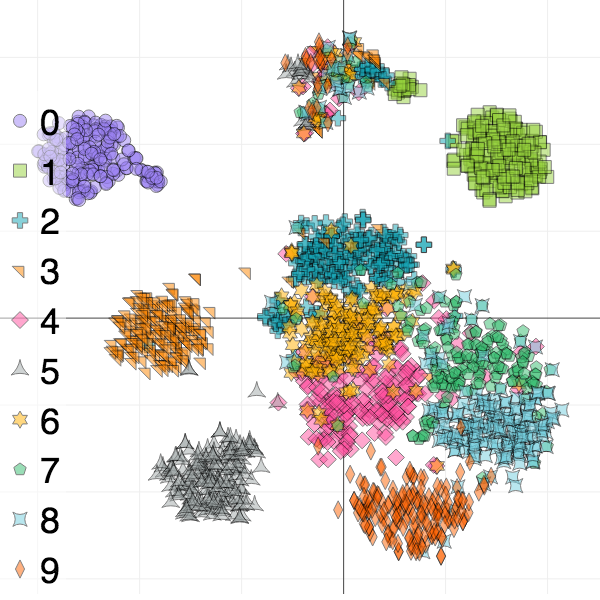}
\caption{Sign-lang. (IM)}\label{embedding-plots-sign-1}
\end{subfigure}\\%
\begin{subfigure}[b]{0.199\textwidth}
\includegraphics[width=\textwidth]{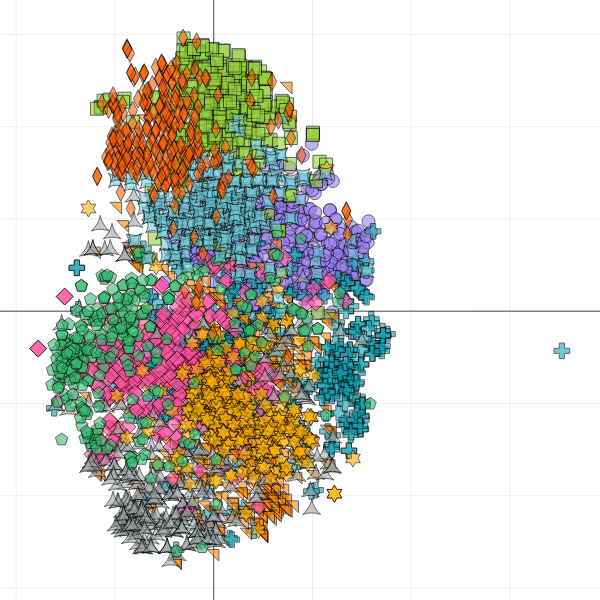}
\caption{CIFAR-10 (TF)}\label{embedding-plots-cifar-2}
\end{subfigure}%
\begin{subfigure}[b]{0.199\textwidth}
\includegraphics[width=\textwidth]{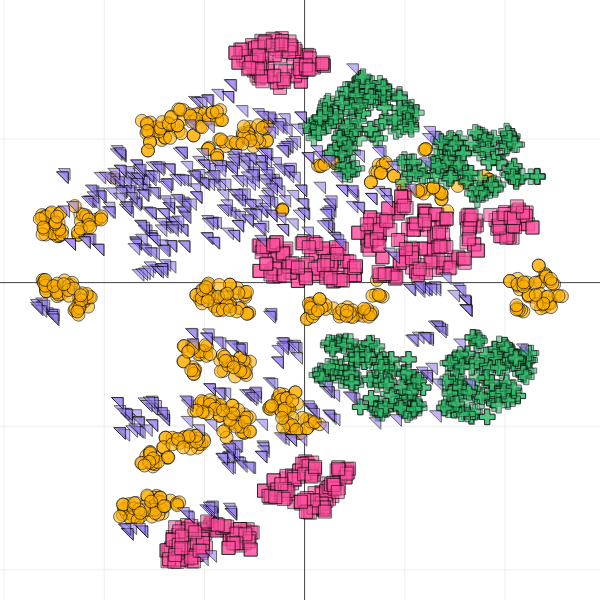}
\caption{Blood cells (TF)}\label{embedding-plots-blood-2}
\end{subfigure}%
\begin{subfigure}[b]{0.199\textwidth}
\includegraphics[width=\textwidth]{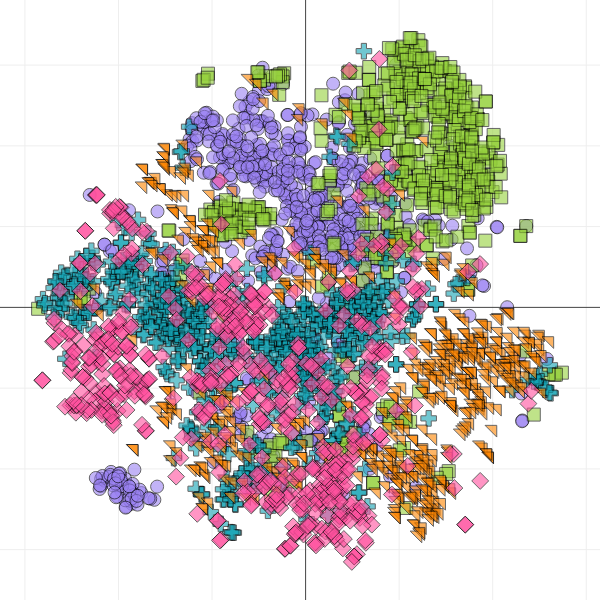}
\caption{Flowers (TF)}\label{embedding-plots-flowers-2}
\end{subfigure}%
\begin{subfigure}[b]{0.199\textwidth}
\includegraphics[width=\textwidth]{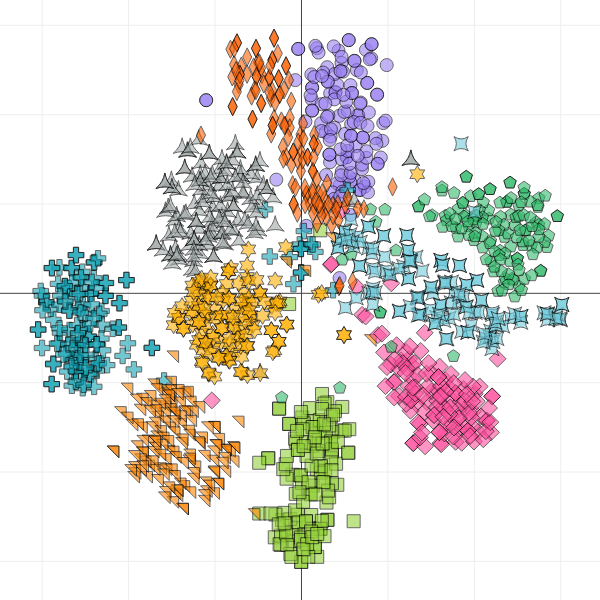}
\caption{Monkeys (TF)}\label{embedding-plots-monkey-2}
\end{subfigure}%
\begin{subfigure}[b]{0.199\textwidth}
\includegraphics[width=\textwidth]{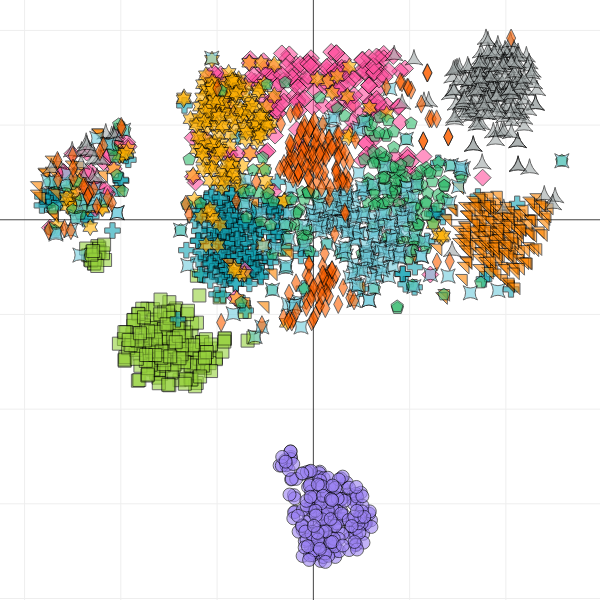}
\caption{Sign-lang. (TF)}\label{embedding-plots-sign-2}
\end{subfigure}%
\caption{3D t-SNE of ImageNet (IM) and Transfer-Frozen (TF) features.}
\label{embedding-plots-fig}
\end{figure}

\textbf{Embedding Quality}: k-nearest neighbor (kNN) accuracy in the resultant embedding measures local homogeneity, which can have positive correlation with labeling efficiency given that more homogeneous clusters can be labeled with fewer corrective actions.
Table~\ref{tab-inc-feats} compares the kNN accuracies, where transfer learning significantly improves accuracy for blood cells, flowers, and monkeys.
The assumption with rotation-based Exemplar-CNN is that the data will have rotation-invariant class interpretation, which may not hold so well for sign-language.
Unsupervised t-SNE plots are shown in Figure~\ref{embedding-plots-fig} of the different features for the datasets, with the expectation of improved clustering with transfer learning.
ImageNet-pretrained features already provide good embeddings for CIFAR-10, flowers, and sign-language, possibly because of semantic overlap. Blood cells and monkeys appear to benefit considerably from transfer learning, even if the CNN stage is frozen.


\section{Improved Semi-Supervised t-SNE}

\textbf{Semi-supervised t-SNE}: t-SNE performs a probabilistic reconstruction of high-dimensional pairwise similarities $p_{ij}$ in an embedding space to produce low-dimensional similarities $q_{ij}$.
Student-t kernel $t_{ij}=(1+||\mathbf{y}_i-\mathbf{y}_j||^2)^{-1}$ measures similarity between embedding points $\mathbf{y}_i$ and $\mathbf{y}_j$, which are normalized as $q_{ij}=t_{ij}/\sum_{k\neq l}t_{kl}$.
The Kullback-Leibler divergence $C=\sum_{i\neq j}p_{ij}\log(p_{ij}/q_{ij})$ between the similarity distributions is iteratively optimized with gradient descent.
The t-SNE gradient can be decomposed into attractive and repulsive forces $\frac{\partial C}{\partial \mathbf{y}_i} = 4(F_{\text{attr}}-F_{\text{rep}})$ as in (\ref{van2013barnes2}) so that an approximation of this $\mathcal{O}(N^2)$ algorithm can be devised~\cite{van2013barnes}. This decomposition can be used to affect pairwise constraints $a_{ij}$ and $b_{ij}$ between labeled points, such as promotion of attraction between same-class samples or repulsion for a label difference.
\begin{equation}\label{van2013barnes2}\small
\frac{\partial C}{\partial \mathbf{y}_i} = 4\sum_{i\neq j}\left(\frac{a_{ij}p_{ij}t_{ij}(\mathbf{y}_i-\mathbf{y}_j)}{\sum_{k\neq l}a_{kl}p_{kl}}-\frac{b_{ij}t_{ij}^2(\mathbf{y}_i-\mathbf{y}_j)}{\sum_{k\neq l}b_{kl}t_{kl}}\right)
\end{equation}

\textbf{Attraction and Repulsion}: The Bayesian attractive priors $a_{ij}$ of~\cite{mcinnes2016sstsne} effectively modifies the high-dimensional similarity $p_{ij}$ according to label differences and labeling importance $f\ge 0$ and point learning rates $u_i$, which are 0 until annotation $c_i$ is assigned with a gradual rate increase to 1 to compensate for exaggerated forces after simultaneous labeling.
$a_{ij}=1/N$ unless both $i$ and $j$ are labeled then if $c_i=c_j$ use $a_{ij}=1/N+u_iu_jf/N_s$, or if $c_i\neq c_j$ use $a_{ij}=1/N-u_iu_jf/N_o$. Here $N_s$ and $N_o$ is the number of same-label and different-label samples to $c_i$, respectively.
A repulsion-emphasizing scalar $r\ge 0$ upscales the repulsive force between two points with mismatched labels so that $b_{ij}=1+u_iu_jr$, otherwise $b_{ij}=1$. The hypothesis is that repulsion emphasis likely creates wider cluster boundaries, which improves cluster homogeneity.

\begin{figure}[t!]%
\centering%
\begin{subfigure}[b]{0.44\textwidth}
\includegraphics[width=\textwidth]{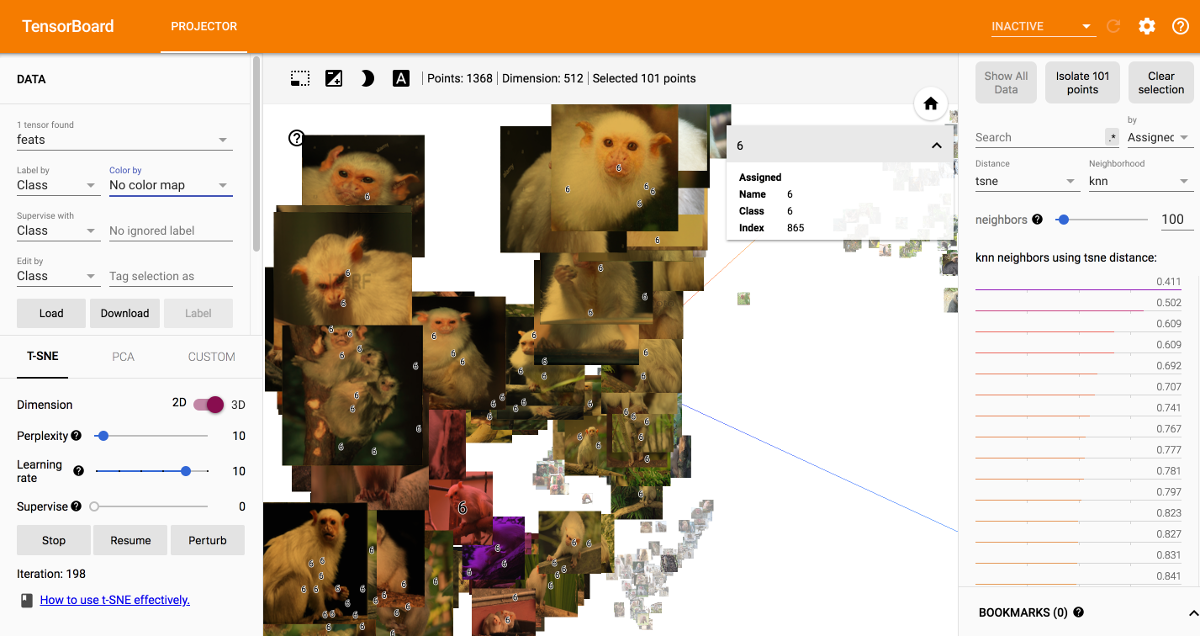}
\caption{TensorBoard Projector}\label{monkey-tensorboard-1}
\end{subfigure}
\begin{subfigure}[b]{0.26\textwidth}
\includegraphics[width=\textwidth]{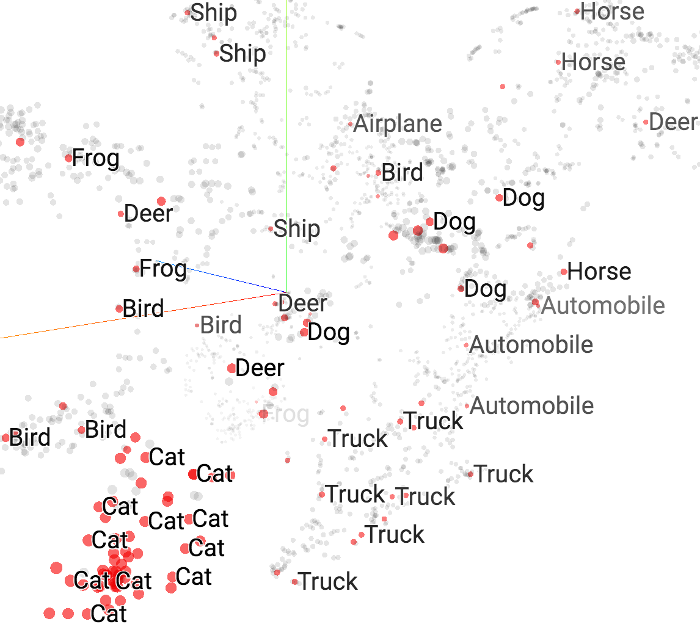}
\caption{Cat sample}\label{bh-kernel-1}
\end{subfigure}%
\begin{subfigure}[b]{0.26\textwidth}
\includegraphics[width=\textwidth]{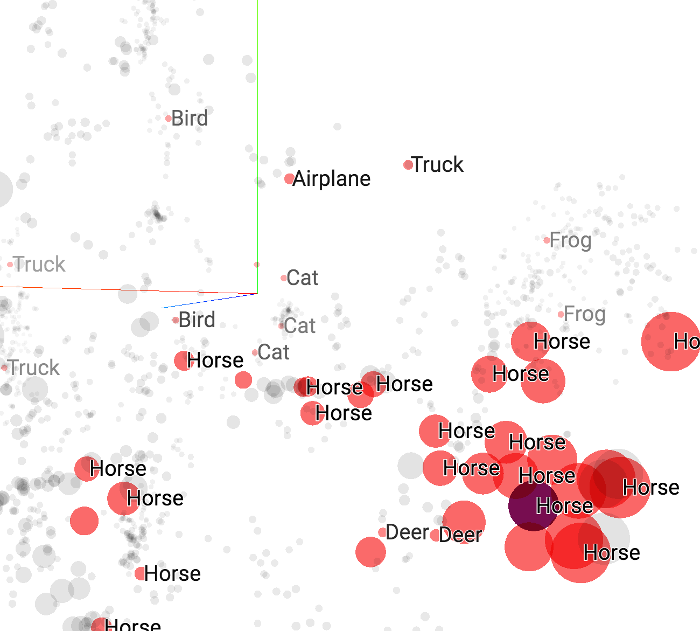}
\caption{Horse sample}\label{bh-kernel-2}
\end{subfigure}%
\caption{(a) TensorBoard Projector~\cite{smilkov2016embedding} with monkey dataset showing a homogeneous cluster selection. (b, c) Barnes-Hut neighbors in 3D t-SNE of CIFAR-10.}\label{bh-kernel-fig}
\end{figure}

\textbf{Barnes-Hut-SNE}: The Barnes-Hut-SNE implementation of t-SNE is used as it reduces complexity to $\mathcal{O}(N\log N)$ by calculating pairwise repulsive forces only for close neighborhoods, while approximating distant clusters with partition trees as if it is a single point~\cite{van2013barnes}. Distant clusters relative to a given embedding point are summarized by utilizing its cell diameter $d_{\text{cell}}$ and center-of-mass $\mathbf{y}_{\text{cell}}$ in the distance calculation.
Octrees are used, which are partition trees providing fast $\mathcal{O}(N)$ construction and neighbor querying, which decompose space into adaptable cells that split when maximum capacity is reached~\cite{finkel1974quad}.
Repulsive forces on an embedding point are calculated where the octree is traversed from the root node with cells nearby the point entered and distant cells pruned and summarized. 
Cells are entered in the Barnes-Hut traversal for $i$ when $d_{\text{cell}}\left/\|\mathbf{y}_i-\mathbf{y}_{\text{cell}}\|\right.>\theta$, which then form part of an approximate kNN($i$) where the neighborhood density decreases rapidly with relative distance. This kNN selection process is used as a focus kernel to model the annotator's local homogeneity assessment at each point, with examples from a CIFAR-10 t-SNE shown in Figure~\ref{bh-kernel-fig}.
\begin{figure}[b!]
\centering
\begin{subfigure}[b]{0.157\textwidth}
\includegraphics[width=\textwidth]{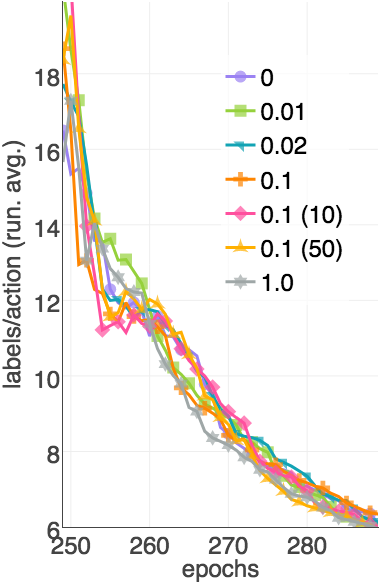}
\caption{$f$,$r$=0,$s$=250}\label{label-importance-3}
\end{subfigure}%
\begin{subfigure}[b]{0.162\textwidth}
\includegraphics[width=\textwidth]{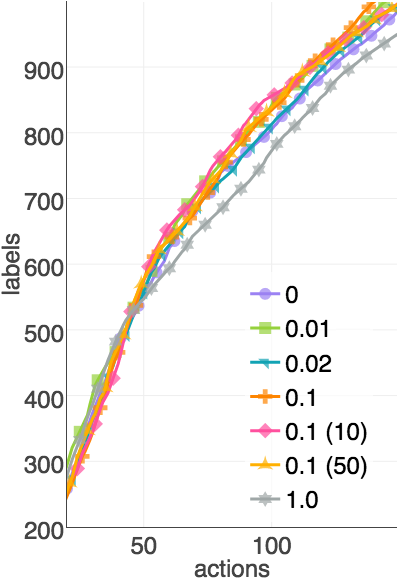}
\caption{$f$,$r$=0,$s$=250}\label{label-importance-1}
\end{subfigure}%
\begin{subfigure}[b]{0.16\textwidth}
\includegraphics[width=\textwidth]{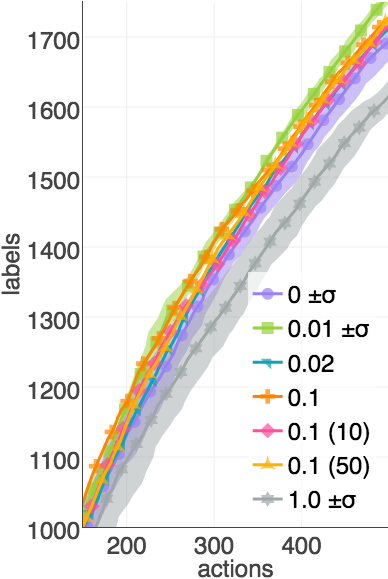}
\caption{$f$,$r$=0,$s$=250}\label{label-importance-2}
\end{subfigure}
\begin{subfigure}[b]{0.158\textwidth}
\includegraphics[width=\textwidth]{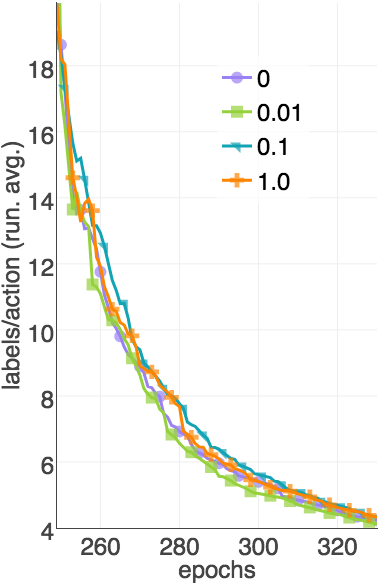}
\caption{$r$,$s$=250}\label{semi-sup-opt-1}
\end{subfigure}%
\begin{subfigure}[b]{0.165\textwidth}
\includegraphics[width=\textwidth]{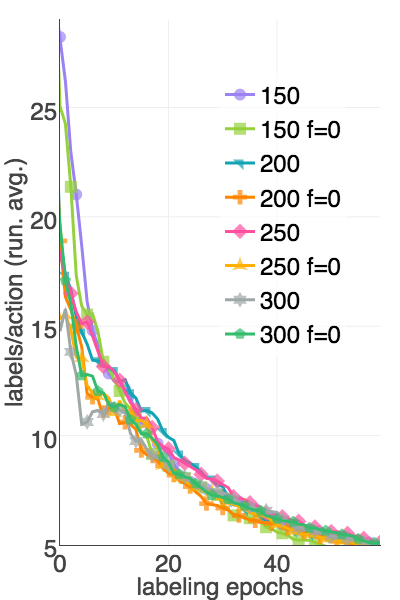}
\caption{$s$,$r$=0.1}\label{semi-sup-opt-3}
\end{subfigure}%
\begin{subfigure}[b]{0.157\textwidth}
\includegraphics[width=\textwidth]{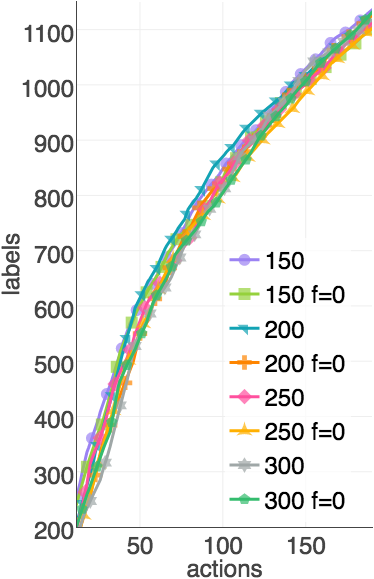}
\caption{$s$,$r$=0.1}\label{semi-sup-opt-2}
\end{subfigure}\\%
\begin{subfigure}[b]{0.137\textwidth}
\includegraphics[width=\textwidth]{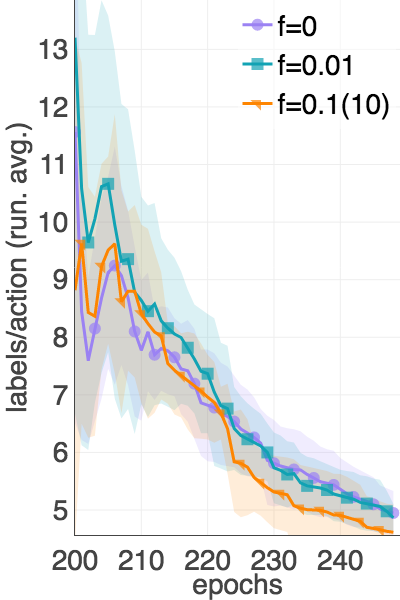}
\caption{CIFAR-10}\label{eff-vs-epochs-cifar}
\end{subfigure}%
\begin{subfigure}[b]{0.119\textwidth}
\includegraphics[width=\textwidth]{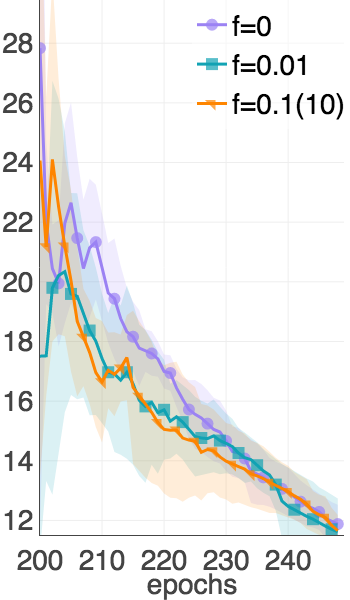}
\caption{Blood}\label{eff-vs-epochs-blood}
\end{subfigure}%
\begin{subfigure}[b]{0.119\textwidth}
\includegraphics[width=\textwidth]{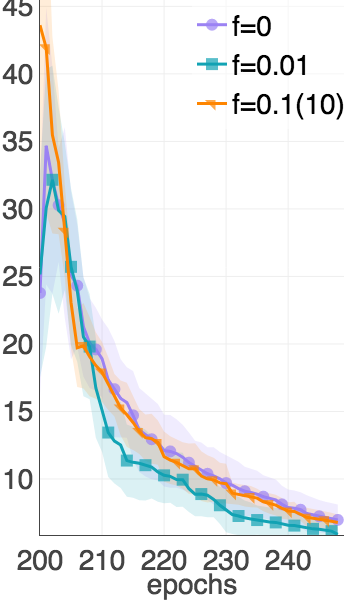}
\caption{Flowers}\label{eff-vs-epochs-flowers}
\end{subfigure}%
\begin{subfigure}[b]{0.119\textwidth}
\includegraphics[width=\textwidth]{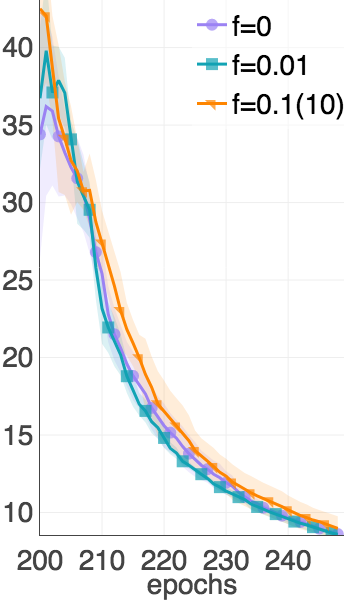}
\caption{Monkeys}\label{eff-vs-epochs-monkey}
\end{subfigure}
\begin{subfigure}[b]{0.137\textwidth}
\includegraphics[width=\textwidth]{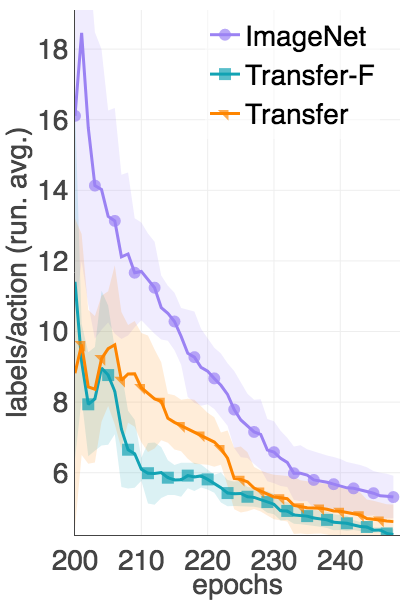}
\caption{CIFAR-10}\label{transfer-eff-vs-epochs-cifar}
\end{subfigure}%
\begin{subfigure}[b]{0.119\textwidth}
\includegraphics[width=\textwidth]{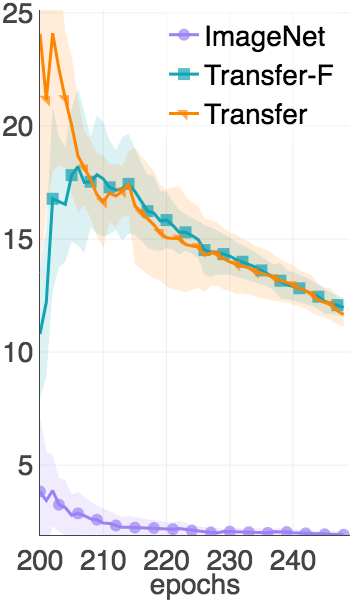}
\caption{Blood}\label{transfer-eff-vs-epochs-blood}
\end{subfigure}%
\begin{subfigure}[b]{0.119\textwidth}
\includegraphics[width=\textwidth]{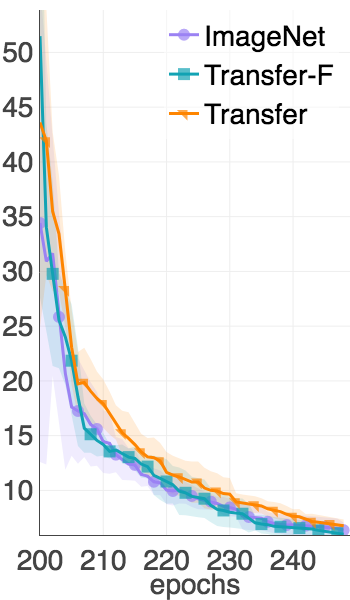}
\caption{Flowers}\label{transfer-eff-vs-epochs-flowers}
\end{subfigure}%
\begin{subfigure}[b]{0.119\textwidth}
\includegraphics[width=\textwidth]{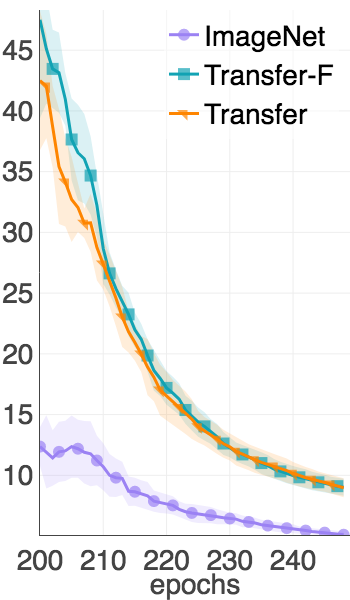}
\caption{Monkeys}\label{transfer-eff-vs-epochs-monkey}
\end{subfigure}%
\caption{(a-c) Semi-supervised t-SNE labeling importance comparison for CIFAR-10 (ImageNet). (d-f) Semi-supervised t-SNE repulsion and starting epoch comparison ($f$=0.01) for CIFAR-10 (IM).
(g-j) Labeling efficiency comparison with fully transferred (TF) features ($r$=0.1 for $f>$0, and $s$=200). (k-n) Labeling efficiency comparison of transfer learning with t-SNE ($f$=0.1(10), $r$=0.1, $s$=200).}
\label{label-importance-fig}
\end{figure}

\textbf{t-SNE Parameters}: The Barnes-Hut-SNE implementation and parameter settings are followed with a few exceptions to facilitate interactive labeling by smoothing early parameter settings in the formative period typically between epochs 100-250. Linear scheduling for local optima parameter $\alpha$ is set decrementally from 4 at epoch 100 to 1 at epoch 250, and for the additional momentum term from 0.5 to 0.8.
The effective neighborhood size or perplexity is fixed at 20, which is empirically adequate for the considered datasets.
The normal Barnes-Hut-SNE approximation is done with $\theta$=0.5, but $\theta$=0.8 is used for tighter Barnes-Hut neighborhoods.
All experimental results are performed across five training subsets in a five-fold cross-validation partitioning of the given data.
Labeling importance $f$ is optimized on CIFAR-10 in Figure~\ref{label-importance-fig} for a fixed starting epoch $s$=250 and no repulsion emphasis, using either an immediate point learning rate $u_i$=1, or linear schedules over 10 or 50 epochs in the case of 0.1 (10) and 0.1 (50), respectively. Larger factors $f>0.1$ typically appear to reduce labeling efficiency, and gradual point learning rate schedules are then required to outperform unsupervised t-SNE.
Fixing labeling importance to $f$=0.01, the effect of repulsion emphasis is inspected in Figure~\ref{semi-sup-opt-1}, with the efficiency increase with a positive emphasis $r$=0.1 seen over a broad range of labeling epochs.
Fixing both $f$=0.01 and $r$=0.1 a variety of starting epochs are investigated in Figure~\ref{semi-sup-opt-2}~and~\ref{semi-sup-opt-3}, with $s$=200 showing greater efficiency in the mid-range of labeling epochs. This corresponds to the last formation phase with a higher learning rate, which hypothetically assists in escaping local optima.



\begin{algorithm}[htpb!]
\caption{t-SNE labeling simulation with annotation emulation using Barnes-Hut neighborhoods.\label{algo-1}}
\small
\SetAlgoLined\SetArgSty{}
\KwData{Features $\mathbf{x}$, true labels $l$ (assess homogeneity), partial labels $c$ (annotated), and t-SNE process $S_e(\mathbf{x},l)$}
\KwResult{Yield updated labels $c$ obtained through singular/groupwise annotation in the labeling interface}
\While(\tcp*[f]{current epoch e after s (labeling start) until stopping epoch}){$s\le e \le e_{\text{max}}$}{
    $\mathbf{y}\leftarrow S_e(\mathbf{x},c)$\tcp*[f]{update t-SNE low-dim positions conditioned on annotated labels c}

    $T\leftarrow\text{octtree}(\mathbf{y})$\tcp*[f]{construct low-dim octtree for BH-neighborhood determination}

    \For(\tcp*[f]{for every point do approx count of labeling opportunities}){$i\in[0,N)$}{
        $n_i\leftarrow 0$\tcp*[f]{to count same-class unlabeled (c) BH-NN, compare classes with l}
        
        $M\leftarrow[\text{root}(T)],\quad m\leftarrow 0$\tcp*[f]{init cell array with tree root to search for BH-NN}

        \While(\tcp*[f]{traverse tree by entering cells nearby i into M}){$m<|M|$}{
            $z\leftarrow M[m],\quad m\leftarrow m+1$\tcp*[f]{get next cell to check distance ratio to i}

            \If(\tcp*[f]{check if nearby using cell diameter and centroid}){$d_z\left/\|\mathbf{y}_i-\mathbf{y}_z\|\right.>\theta_{K}$}{
                \If(\tcp*[f]{is anchor sample of cell still unlabeled?}){$c_z=\oslash$}{
                    \If(\tcp*[f]{check if same class with true labels l (homogeneity)}){$l_z=l_i$}{
                        $n_i\leftarrow n_i+1$\tcp*[f]{add groupwise labeling opportunity to focus i}
                    }
                }
                $M\leftarrow [M, \text{children}(z)]$\tcp*[f]{traverse cell children or terminate leaf}
            }
        }
    }
    $v\leftarrow \arg\max_in_i$\tcp*[f]{focus sample with approx most obtainable groupwise labels}

    $W(k)\leftarrow \text{sorted}\left(\text{kNN}(v)\right)$\tcp*[f]{exact kNN of focus sample v, sorted ascending distance}

    $\text{actions}\leftarrow 1,\quad \text{labels}\leftarrow 1$\tcp*[f]{record 1 action to select v and get its label}

    \For(\tcp*[f]{from closest to furthest kNN (emulate size slider)}){$k\in[1,|W|],\quad j\in W(k)$}{
        \If(\tcp*[f]{now groupwise selection and not just focus sample}){$k=1$}{
        $\text{actions}\leftarrow \text{actions}+1$\tcp*[f]{once-off 1 action to set kNN size slider to optimum k}
        }
        \If(\tcp*[f]{unlabeled (note that labeled samples keep unchanged)}){$c_j=\oslash$}{
            \If(\tcp*[f]{visual local homogeneity assessment of same class}){$l_j=l_i$}{
                $\text{labels}\leftarrow \text{labels}+1$\tcp*[f]{label j as part of groupwise labeling all same label}
            }
            \Else(\tcp*[f]{otherwise deselect sample from groupwise selection to avoid error}){
                $\text{actions}\leftarrow \text{actions}+1$\tcp*[f]{always count 1 action to deselect differing j}
            }
        }
        $h_k\leftarrow \text{labels}/\text{actions}$\tcp*[f]{groupwise labeling efficiency up to kNN to find best k}
    }
    \For(\tcp*[f]{for every NN in most efficient kNN do groupwise}){$j\in W(\arg\max_kh_k)$}{
        \If(\tcp*[f]{unlabeled, no overwriting of existing annotations}){$c_j=\oslash$}{
            \If(\tcp*[f]{same class for groupwise execution and low-action count}){$l_j=l_i$}{
                $c_j=l_i$\tcp*[f]{annotate to assign label j in c from true labels l}
            }
        }
    }
}
\end{algorithm}

\textbf{Labeling Efficiency}: 
The efficiency is compared for other datasets with fully transferred features in Figure~\ref{label-importance-fig}. Semi-supervision assists in the case of CIFAR-10 (slight label importance) and monkeys (stronger label importance). However, there is typically a significant overlap in the standard deviation bands of the results, which indicates very similar performance overall.
The expectation is thus that transferred features will provide for greater labeling efficiency, which is seen in particular for blood cells and monkeys. However, with CIFAR-10 the unsupervised transfer learning likely deteriorates already-good ImageNet features given the overlap between the datasets.


\section{Active Learning with t-SNE}
Labeling with t-SNE is an active learning approach driven through local homogeneity visual assessment to find efficient prospects. A comparison is made to uncertainty sampling by continually retraining a classifier with obtained labels, despite not utilizing classifier uncertainty in the t-SNE labeling.
A neural network classifier is periodically trained with all available labeled samples and the SoftMax prediction vector is used to decide which samples to label next, where each point labeling constitutes one action.
The baseline comparison for active learning performance is uncertainty sampling variants including maximum uncertainty $\max_lp(c_i=l|\mathbf{x}_i)$, minimum margin $\max_lp(c_i=l|\mathbf{x}_i)-\max_{m\setminus l}p(c_i=m|\mathbf{x}_i)$, and maximum entropy $-\sum_l p(c_i=l|\mathbf{x}_i)\log p(c_i=l|\mathbf{x}_i)$.
\begin{table*}[t!]
\caption{Average number of actions to reach 80\% of full labeling accuracy, comparing between IM, TF and TR features, and between active learning with uncertainty sampling and t-SNE.}
\label{tab-eff-stats}
\centering
{\small
\begin{tabular}{@{ }l@{} @{ }r@{ } @{ }r@{ } @{ }r@{ } @{ }r@{ } @{ }r@{ } @{ }r@{ } @{ }r@{ } @{ }r@{ } @{ }r@{ } @{ }r@{ } @{ }r@{ } @{ }r@{ } @{ }r@{ } @{ }r@{ } @{ }r@{ } @{ }r@{ } @{ }r@{ } @{ }r@{ } @{ }r@{ } @{ }r@{ } @{ }r@{ }}
\hline
&\multicolumn{3}{c}{\bf{CIFAR-10}}&&\multicolumn{3}{c}{\bf{Blood cells}}&&\multicolumn{3}{c}{\bf{Flowers}}&&\multicolumn{3}{c}{\bf{Monkeys}}&&\multicolumn{3}{c}{\bf{Sign-lang.}}&&	\\
\cline{2-4} \cline{6-8} \cline{10-12} \cline{14-16} \cline{18-20}
&	IM	&	TF	&	TR	&	&	IM	&	TF	&	TR	&	&	IM	&	TF	&	TR	&	&	IM	&	TF	&	TR	&	&	IM	&	TF	&	TR	&&	Avg.	\\
\hline
RandSelect 	&	170	&	120	&	130	&	&	180	&	120	&	150	&	&	130	&	70	&	50	&	&	200	&	60	&	60	&	&	140	&	100	&	90	&&	118	\\
Uncertainty 	&	240	&	150	&	150	&	&	270	&	140	&	130	&	&	290	&	70	&	80	&	&	210	&	40	&	50	&	&	180	&	90	&	80	&&	145	\\
Margin 	&	210	&	120	&	110	&	&	270	&	110	&	120	&	&	220	&	80	&	60	&	&	200	&	40	&	40	&	&	150	&	70	&	70	&&	125	\\
Entropy 	&	240	&	190	&	190	&	&	250	&	140	&	120	&	&	310	&	90	&	70	&	&	210	&	50	&	50	&	&	180	&	90	&	80	&&	151	\\
t-SNE f=0 	&	68	&	86	&	77	&	&	247	&	78	&	\bf{58}	&	&	59	&	37	&	36	&	&	67	&	18	&	27	&	&	37	&	\bf{37}	&	49	&&	\bf{65}	\\
t-SNE f=0.01 	&	\bf{67}	&	\bf{86}	&	\bf{68}	&	&	\bf{247}	&	83	&	68	&	&	72	&	\bf{36}	&	\bf{26}	&	&	\bf{66}	&	\bf{18}	&	\bf{26}	&	&	\bf{37}	&	57	&	\bf{47}	&&	67	\\
t-SNE f=0.1(10) 	&	68	&	106	&	77	&	&	248	&	\bf{68}	&	68	&	&	\bf{56}	&	37	&	27	&	&	68	&	27	&	28	&	&	37	&	62	&	49	&&	68	\\
\hline
Average	&	152	&	123	&	\bf{115}	&	&	245	&	106	&	\bf{102}	&	&	162	&	60	&	\bf{50}	&	&	146	&	\bf{36}	&	40	&	&	109	&	72	&	\bf{66}	&&		\\
\hline
\end{tabular}
}
\end{table*}
\begin{figure}[b!]%
\centering%
\begin{subfigure}[b]{0.1338\textwidth}
    \includegraphics[width=\textwidth]{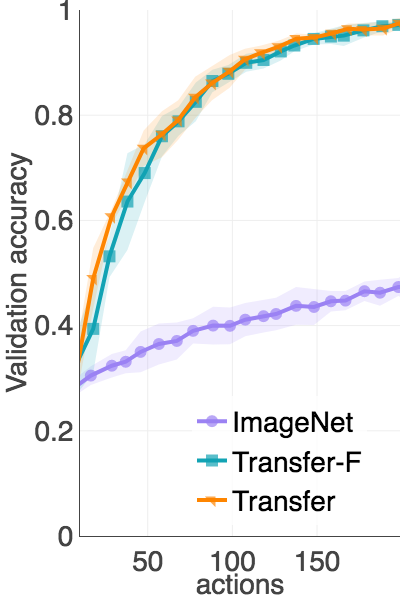}
    \caption{Blood cells}\label{al-act-vs-val-blood-4}
\end{subfigure}%
\begin{subfigure}[b]{0.119\textwidth}
    \includegraphics[width=\textwidth]{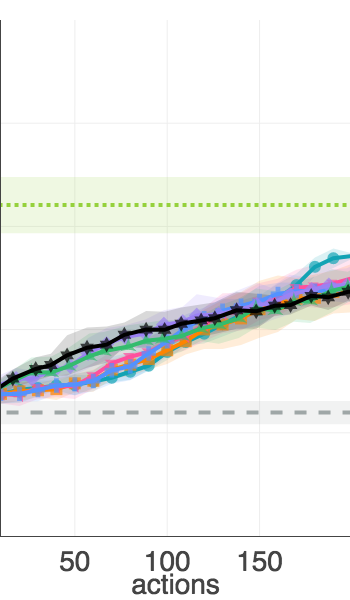}
    \caption{IM}\label{al-act-vs-val-blood-1}
\end{subfigure}%
\begin{subfigure}[b]{0.119\textwidth}
    \includegraphics[width=\textwidth]{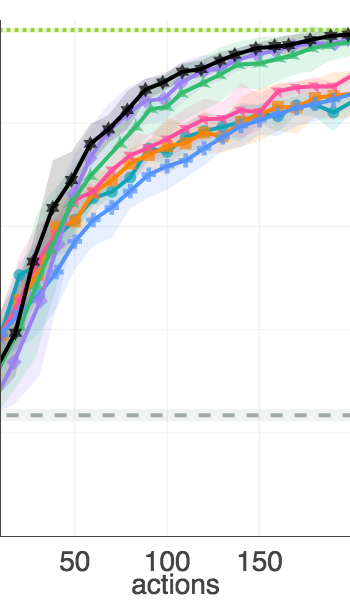}
    \caption{TF}\label{al-act-vs-val-blood-2}
\end{subfigure}%
\begin{subfigure}[b]{0.119\textwidth}
    \includegraphics[width=\textwidth]{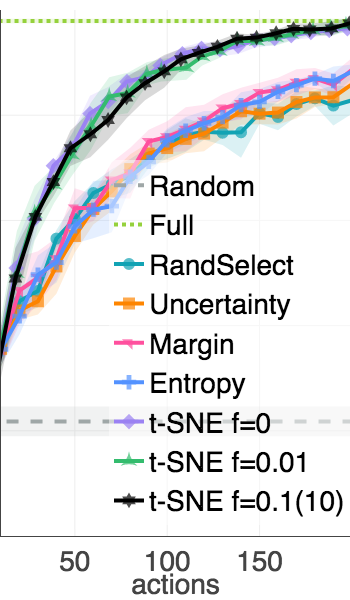}
    \caption{TR}\label{al-act-vs-val-blood-3}
\end{subfigure}%
\begin{subfigure}[b]{0.1338\textwidth}
    \includegraphics[width=\textwidth]{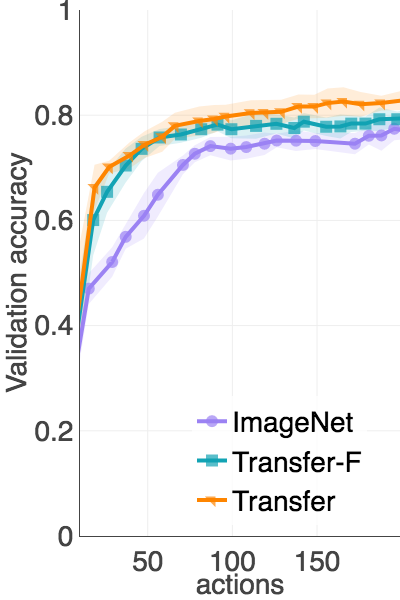}
    \caption{Flowers}\label{al-act-vs-val-flowers-4}
\end{subfigure}%
\begin{subfigure}[b]{0.119\textwidth}
    \includegraphics[width=\textwidth]{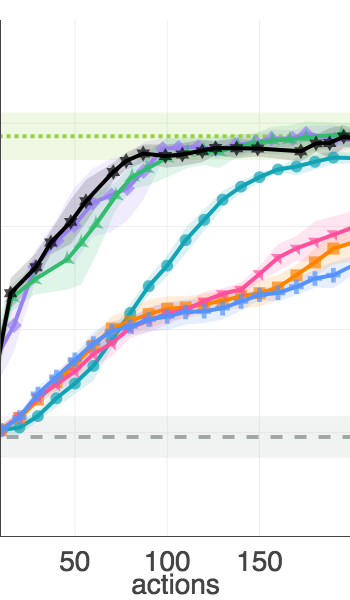}
    \caption{IM}\label{al-act-vs-val-flowers-1}
\end{subfigure}%
\begin{subfigure}[b]{0.119\textwidth}
    \includegraphics[width=\textwidth]{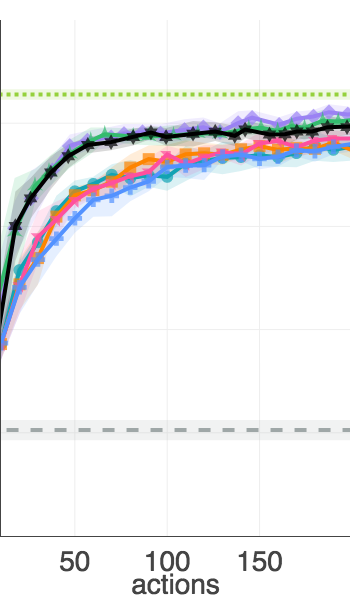}
    \caption{TF}\label{al-act-vs-val-flowers-2}
\end{subfigure}%
\begin{subfigure}[b]{0.119\textwidth}
    \includegraphics[width=\textwidth]{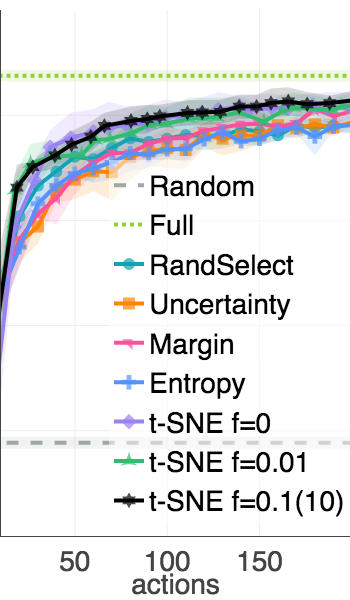}
    \caption{TR}\label{al-act-vs-val-flowers-3}
\end{subfigure}\\%
\begin{subfigure}[b]{0.1338\textwidth}
    \includegraphics[width=\textwidth]{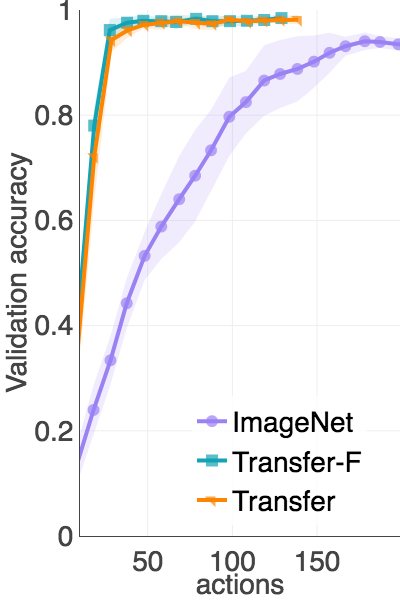}
    \caption{Monkeys}\label{al-act-vs-val-monkey-4}
\end{subfigure}%
\begin{subfigure}[b]{0.119\textwidth}
    \includegraphics[width=\textwidth]{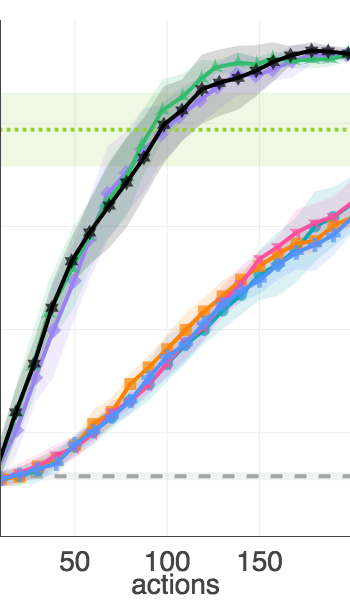}
    \caption{IM}\label{al-act-vs-val-monkey-1}
\end{subfigure}%
\begin{subfigure}[b]{0.119\textwidth}
    \includegraphics[width=\textwidth]{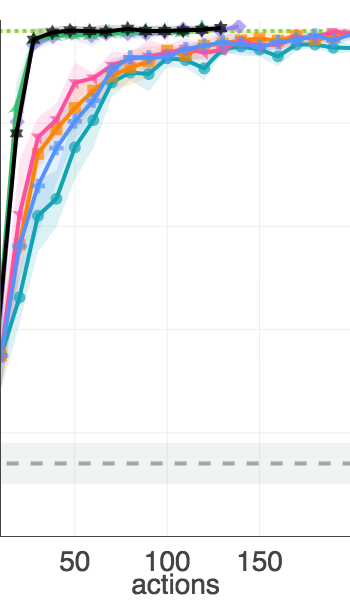}
    \caption{TF}\label{al-act-vs-val-monkey-2}
\end{subfigure}%
\begin{subfigure}[b]{0.119\textwidth}
    \includegraphics[width=\textwidth]{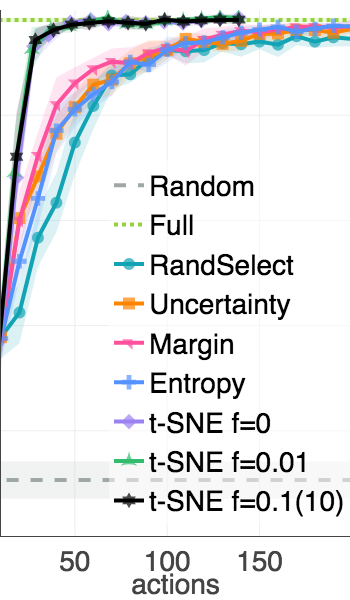}
    \caption{TR}\label{al-act-vs-val-monkey-3}
\end{subfigure}%
\begin{subfigure}[b]{0.1338\textwidth}
    \includegraphics[width=\textwidth]{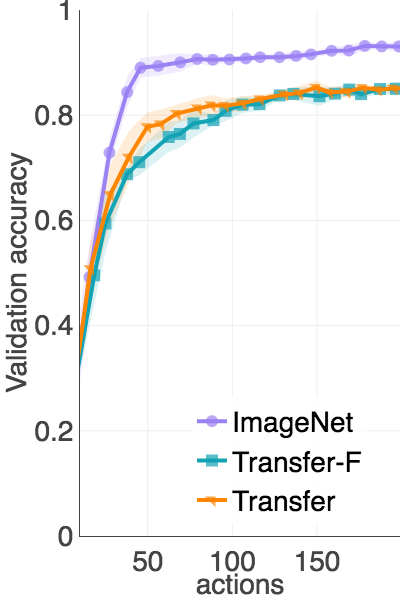}
    \caption{Sign-lang.}\label{al-act-vs-val-sign-4}
\end{subfigure}%
\begin{subfigure}[b]{0.119\textwidth}
    \includegraphics[width=\textwidth]{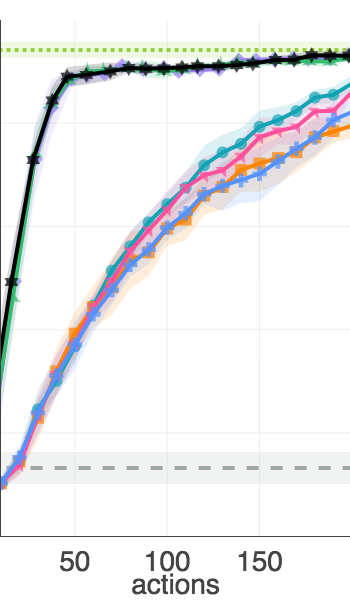}
    \caption{IM}\label{al-act-vs-val-sign-1}
\end{subfigure}%
\begin{subfigure}[b]{0.119\textwidth}
    \includegraphics[width=\textwidth]{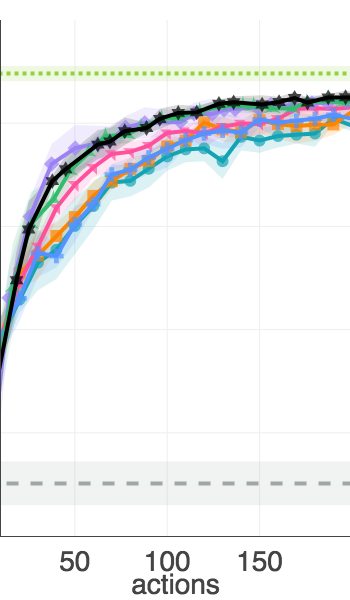}
    \caption{TF}\label{al-act-vs-val-sign-2}
\end{subfigure}%
\begin{subfigure}[b]{0.119\textwidth}
    \includegraphics[width=\textwidth]{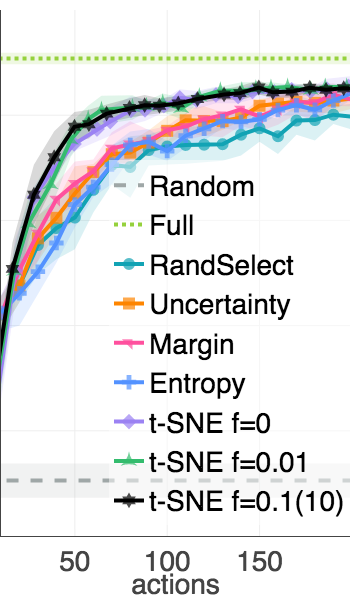}
    \caption{TR}\label{al-act-vs-val-sign-3}
\end{subfigure}%
\caption{Active learning comparison between uncertainty sampling and t-SNE labeling ($f$=0.1(10), $r$=0.1, and $s$=200) for blood cells (a-d), flowers (e-h), monkeys (i-l), and sign-language (m-p).}
\label{al-act-vs-val-fig}
\end{figure}
Five-fold cross-validation is performed on each dataset with a neural network consisting of an input layer ($D$ nodes), 25\% dropout, a ReLU-activated dense hidden layer ($D/4$ nodes), 25\% dropout, and a SoftMax classification layer. The network is trained from its existing state with available labels for 50 training epochs, before 10 new points are sampled for labeling, in the case of uncertainty sampling and random selection (RandSelect). The network is continually retrained as new labels are accumulated. t-SNE labels are used in a similar manner with retraining in the closest increments of 10 actions, since the number of actions per prospect varies.
Validation accuracy is measured against actions in Figure~\ref{al-act-vs-val-fig}, and unsupervised and semi-supervised t-SNE show very similar performance. Exceptions include CIFAR-10 (Transfer-F) where greater semi-supervision performs worse, and flowers (ImageNet) where greater semi-supervision performs better.
t-SNE is seen to produce greater accuracy for significantly fewer labeling actions than uncertainty sampling. Table~\ref{tab-eff-stats} indicates that random selection performs relatively well, possibly due to the small dataset sizes. The gap between t-SNE and uncertainty sampling seems to decrease for higher quality features.
In Figures~\ref{al-act-vs-val-blood-4},~\ref{al-act-vs-val-flowers-4}, and ~\ref{al-act-vs-val-monkey-4} it is seen that transferred features (Transfer-F and Transfer) significantly outperform ImageNet features. Across all methods, transferred features display higher validation accuracy for fewer actions.

\section{Conclusion}
Transfer learning with exemplar prediction from a pretrained CNN is proposed for small specialized image datasets in this study, and significant improvements in embedding quality, labeling efficiency, and active learning is observed.
Improved semi-supervision is proposed as a means of interactively labeling an embedding, although the majority of the labeling efficiency is obtained even in the unsupervised case, through abundance of combined labeling prospects in a quality embedding.
t-SNE labeling typically significantly outperforms uncertainty sampling, possibly because of the larger data requirement of uncertainty sampling.
    
\bibliographystyle{unsrt}
\bibliography{refs}







\end{document}